\documentclass[doc,apacite]{apa6}
\usepackage{graphicx}
\usepackage{apacite}
\usepackage{times}
\usepackage{latexsym}
\usepackage{amsmath} 
\usepackage{multirow}
\usepackage{rotating}
\usepackage{url}
\usepackage{subcaption}
\usepackage{algorithm}
\usepackage{algpseudocode}
\usepackage{paralist}
\usepackage{acronym}
\usepackage{mathtools}
\usepackage{helvet}
\usepackage{courier}
\usepackage{amsmath}
\usepackage{multirow} 
\usepackage{enumitem}
\usepackage{algpseudocode}
\usepackage{subcaption}

\title{Multilingual Topic Models}
\shorttitle{}
\fourauthors{Kriste Krstovski}{Michael J. Kurtz}{David A. Smith}{Alberto Accomazzi}
\fouraffiliations{Columbia University\\New York, NY 02138}{Harvard-Smithsonian Center for Astrophysics\\Cambridge, MA 02138}{ Northeastern University\\ Boston, MA 02115}{Harvard-Smithsonian Center for Astrophysics\\Cambridge, MA 02138}

\abstract{Scientific publications have evolved several features for mitigating vocabulary mismatch when indexing, retrieving, and computing similarity between articles.  These mitigation strategies range from simply focusing on high-value article sections, such as titles and abstracts, to assigning keywords, often from controlled vocabularies, either manually or through automatic annotation.  Various document representation schemes possess different cost-benefit tradeoffs.  In this paper, we propose to model different representations of the same article as translations of each other, all generated from a common latent representation in a multilingual topic model.  We start with a methodological overview on latent variable models for parallel document representations that could be used across many information science tasks.  We then show how solving the inference problem of mapping diverse representations into a shared topic space allows us to evaluate representations based on how topically similar they are to the original article.  In addition, our proposed approach provides means to discover where different concept vocabularies require improvement.}

\begin{document}
\maketitle
                        
\section{Introduction}
\label{sec:1}

Designers of information systems for indexing, retrieving, recommending, and computing the similarities of scholarly articles have long contended with the challenge of vocabulary mismatch, as scientific disciplines ramify into different subfields.  In the scholarly literature, moreover, constellations of interrelated technical terms further confound discovery by changing their meaning in concert.  Describing scientific concepts using the vocabulary of one subfield, for the purpose of discovering similar methodologies in another, often fails to provide satisfactory results. 

To expedite the flow of information, scholarly articles often follow a conventional layout, often with standardized section headings. Ideas presented in one section may be interpreted differently and weighted with a different level of significance than another section. For example, the term \textit{Hubble constant} presented in the introductory section of an article may simply be used as a reference in defining a particular measurement, in contrast to the same term presented in a section titled ``Implications for cosmology'' or ``Conclusions''. In addition to the vocabulary mismatch found across different articles, this variation in vocabulary across different sections of an article presents an additional challenge when determining similar articles. 

To address these problems, researchers have often employed representations that abstract away from the actual words used by the authors and that map all articles in a collection into a discrete feature space. Such representations often use features drawn from a controlled vocabulary. Examples of such representations are keyword and concept based representational systems. The goal of these representations is to specify the principal subject matter of the article and to offer a concise conceptual representation. For a given scientific field, in most cases, these vocabularies are generated manually. For example, many scientific publications require authors to provide keyword phrases, i.e. keywords, to their article which the author chooses from a controlled vocabulary specific for their scientific field. 

Similar to keywords, concepts use a controlled vocabulary and they often serve as index terms for performing search and comparison between articles. But unlike keywords which are author based, concepts are generated automatically usually through an elaborate data mining approach that uses different syntactic and semantic extraction techniques and various heuristics. Another distinguished characteristic of concepts is that they are generated from the words found in the document by following collection wide ontology. This is in contrast to keywords specified by authors which do not have to originate from and be anchored with the words in the document. Unlike keywords though, concepts provide a much richer representation which usually contains a larger set of concepts. 

Similar to the text summarization approaches that have been developed in the field of natural language processing (NLP) \cite{Radev_ea_2002}, concept and keyword based representations are often developed with the goal of delivering a summary of the article to the reader. Unlike text summarization approaches, which have a very rich syntactic structure, these types of representations often stand on a much shallower syntactic level. Aside from being a costly and a time consuming process, concept and keyword based annotations, as it is the case with other human processes, are often prone to errors. More importantly they are often explicit of the author bias. 

There exist various keyword and concept based representational approaches which vary based across various dimensions, such as the keywords used, the various heuristics that they employ and the mining approach, to name a few. The plethora of such approaches often makes it difficult to decide which one is most suitable for a collection and especially for the specific task in mind \cite{Hasan_Ng_2014}. In the past, various approaches have been proposed to evaluate different article representations which are task specific and in most cases rely on ground truth annotation. 

In this work we present an evaluation approach that allows us to compare different document representation methodologies using multilingual latent variable models of text such as the multilingual topic models (MLTMs). These models allow us to treat different article representations as translations of the same article in different languages. The underlying assumption of the MLTMs is that documents that are translations of each other, while written in a different language, cover the same set of topics. When represented in the shared topic space document translations reside close to each other versus other documents. In the past this modeling process has been used to detect and retrieve document translation pairs \cite{Mimno_ea_2009}, \cite{Krstovski_Smith_2013}, \cite{Krstovski_Smith_2016}. In our case we extend this modeling approach to map different article representations in a shared topic space using the same underlying assumption -- that topic distributions that are generated from different representations of the same article reside close to each other compared to representations of other articles. 

Our approach is in line with and is driven by the early work of J.R. Firth \cite{Halliday_1971} which defines translation as individual's summarization of the article using one's personal vocabulary of words and one's perception and understanding of the article. Furthermore article's abstract, under this definition, also represents an article translation. 

We compare the similarity of the topical representations of the article derived from the various representational approaches with the topical representation derived from the original article. Assuming that the set of keywords or concepts assigned to the article provide a good article representation one would expect that in this shared topic space different representations of the same article would be positioned close to the topic representation of the original article. 

We demonstrate the efficacy of our approach by evaluating 4 different article representation types which include 2 concept based approaches. We utilized these approaches in order to represent a collection of $\sim$ 32k articles that were published in the Astrophysical Journal (ApJ) over a time period of 10 years. We also show that the number of articles used to train the MTM doesn't significantly affect the efficacy of the evaluation approach especially its ability to rank the different article representations. This makes our approach more practical to use especially when dealing with small article collections such as articles that are published in a fairly new journal. 

While in the past multilingual topic models have been shown to yield good results on tasks of detecting and retrieving document translation pairs to the best of our knowledge, this is the first work that utilizes their modeling approach to evaluate different representation systems. 

An important aspect of the representational approach is its quality across different areas of the collection. A good representation system should provide a uniform representation quality across all articles regardless of their topic. This is especially important when implying a new concept based system and especially during the annotation process when dealing with keyword based representations. Assessing the annotation quality is a time consuming process. And while in some areas of the collection the representation quality could be fairly good there may be instances where the annotation quality could be improved. 

Aside from allowing us to compare different keyword and concept based representational systems, mapping article representations in the shared topic space provides means to highlight representational differences. More specifically, it allows us to derive an approach to evaluate the annotation of journal articles with keywords and concepts. Our evaluation is based on the notion that the assigned keywords and concepts should convey the same or a similar mixture of topics as the actual words in the article. We then look into the articles where the topic distributions across the two representations are not most topically similar. The approach points out articles and determines a particular area or a sub-field where annotation could be improved by observing the assigned keywords or concepts of the topically dissimilar representations.

Multilingual topic models are latent variable models of text. They are a multilingual extension of the latent Dirichlet allocation (LDA) which was originally developed to represent monolingual document collections in the topic space. In order to utilize MLTMs for the purpose of evaluating representation systems we would first need to describe them. We do so by giving an overview of latent variable models of text including LDA followed by a detailed presentation of MLTMs. 

Aside for completeness we believe that giving a detailed description of MLTMs would be beneficial for other information scientists in utilizing MLTMs in other domains and on other tasks. Unlike a recent work by Weng et al. \cite{Wang_ea_2015} which only gives an outline of the Gibbs sampling approach for inferring the posterior distributions in LDA in this paper we give an overview of the Variational Bayes (VB) inference which is more efficient to compute especially its stochastic variant \cite{Krstovski_Smith_2013}. 

We then detail our experimental setup and the keyword and concept based systems that we will be evaluating which is followed by a summary of our experimental results. In the last section we will be presenting our approach for pointing out articles and sub-domains where representational systems could be improved. 

\section{Finding Similar Scholarly Articles}
\label{sec:2}
Assume that we have a collection of scientific articles from various journals in Astrophysics. We have indexed this collection using all article sections, including author-assigned keywords, and have also augmented each article with a set of concepts using various automatic concept representation systems.  Also assume that this collection is part of a large digital library\footnote{For example, \url{http://ads.harvard.edu}} and that we are tasked with developing an article recommender system: given an article that the reader is currently viewing, we would like to display other topically similar articles.  A common approach for solving this similarity-search task is to represent all articles in a shared feature space such as the one that can be generated by keyword or concept based representations. Representing documents in a shared feature space abstracts away from the actual words and their specific sequence in each document and therefore facilitates finding similar articles written using different vocabularies.

Once articles are represented in a shared space, the task of finding similar articles could be viewed as a standard Information Retrieval (IR) task where the collection is indexed and the current viewed document is treated as a query. With the relevance model scoring function we generate a list of documents ranked based on the similarity score. 

\begin{figure*}
  \centering
  \includegraphics[width=0.63\textwidth]{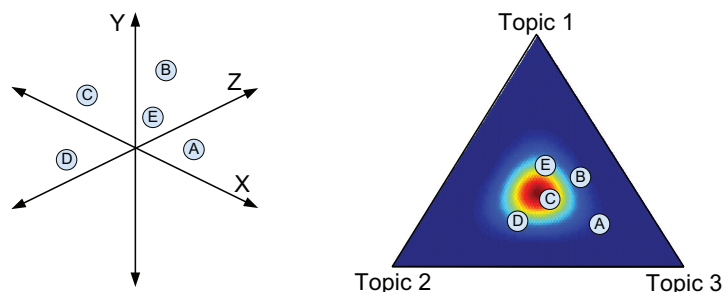}
  \caption{Illustration of articles representation in the metric space (left) and the probability simplex (right). In the metric space articles are represented as feature vectors while in the probability simplex they are represented as probability distributions.}
  \label{fig:metric_simplex}
\end{figure*}

Aside from deciding on the shared features space, another important step in performing document similarity search is deciding on the nature of the shared space and the feature values. Shared space could be either the common metric space or the probability simplex. Figure~\ref{fig:metric_simplex} illustrates a representation of 6 articles in the two shared spaces.

Feature values range from the basic, such as the ones that indicate whether the words, keywords or concepts are present in the document, to collection wide statistics such as term frequency (tf) or inverse document frequency (idf). Vector space models which represent documents in the metric space are exemplar of such a representation. For example, the tf--idf \cite{Croft_ea_2009} model uses a product between the tf and idf features to represent documents in the metric space (i.e. the tf--idf based representation). 

In the past these models have offered good results on various tasks that involve retrieving similar documents. However their drawback is that the generated features are directly related to the words present in the collection and therefore they lack the ability to model documents (or in our case scientific articles) that are presenting work on the same or similar set of topics while using different vocabularies. 

\section{Latent Variable Models of Text}

To overcome the drawbacks of the vector space based representation researchers have proposed models that offer more semantic representation such as the Latent Semantic Indexing (LSI) \cite{Deerwester_ea_1990} which uses singular vector decomposition (SVD) to determine a low-dimensional latent subspace in the space of tf--idf features that better model the variations across the documents in the collection. LSI is based on the principle that "words that are used in the same contexts tend to have similar meanings". With the introduction of statistical modeling techniques a shift towards a better semantic representation was introduced by developing approaches that automatically represent the generative process of the document. In this line of work LSI has been recast as a generative model of text with probabilistic latent semantic indexing (PLSI) \cite{Hofmann_1999}. Unlike LSI, PLSI assigns multinomial distribution over the document which acts like a mixture of topic proportions that is used to draw topic assignment to words in the document. 
While providing better representation than LSI, PLSI has two drawbacks. Firstly it is prone to overfitting as it can only assign probabilities to already seen documents. Secondly, the number of parameters grows linearly with the number of training documents. 

To alleviate these constraints, latent Dirichlet allocation (LDA) \cite{Blei_ea_2003} has been proposed. LDA is a latent variable model of text that uses prior structure over the latent variables to better model the uncertainties in the inferred probability distributions over the hidden variables in the model. PLSI and LDA are representatives of a set of latent variable models of text known as topic models. 

Topic models represent documents as multinomial distributions over a predetermined number of topics. Each topic, on the other hand, is defined as a multinomial distribution over the words in the collection.
This is in contrast to the vector space model and other models whose document representations are anchored with the words used in the document. Furthermore topic based representation is continuous rather than sparse and discrete which is a characteristics of the vector space model. 
In topic models the underlying assumption is that authors goal when generating the document is to convey to the reader a topic or set of topics. Therefore the words are selected from a word distribution for each topic/s. The reader of the document ultimately learns of the author's topics. Therefore topic based representation of documents provides a better semantic representation which is also more compact than the representation using the actual words in the document.

Representing documents in a low-dimensional latent space goes beyond the actual words used in each document and therefore it facilitates deeper semantic analysis of documents written using different vocabularies.
For example, finding similar documents that are written in a highly domain-specific language, such as scientific papers and patents can be very challenging due to the vocabulary differences \cite{Krstovski_ea_2013}. As another example, identifying academic communities that work on related scientific topics can be a challenging task due to the different terminology used across different sub-fields \cite{Talley_ea_2011}. 

In the past feasibility and effectiveness of topic models have been explored across various computer science fields. For example, in IR, Wei \& Croft \cite{Wei_Croft_2006} used LDA to improve document smoothing and ad-hoc retrieval.  In \cite{Blei_ea_2003} authors used LDA on the task of automatic annotation of images. In most cases topic models are used as exploratory tools for performing data analysis such as in Hall et al. \cite{Hall_ea_2008} where authors used LDA to analyze historical trends in scientific literature. 

\subsection{Latent Dirichlet Allocation (LDA)}
\label{sec:2_1}
LDA is the most commonly used topic model. It falls under the category of "admixture" models since it assigns documents with a mixture of topics in contrast to the "mixture" model type where there is an exclusive OR across the possible topics assigned for each document. LDA assumes that the words in the document are the only observable variables and that there is a specific number of topics $K$ in the given collection set a priori. Given a collection of $D$ documents ($d=1,2,...,D$) and a vocabulary of $V$ words, LDA models the generative process of each document in the collection by first generating collection wide topic-word distributions -- for each of the $K$ topics in the collection ($k=1,2,...,K$) it draws a $V$ dimensional multinomial distribution $\varphi_{t}$ from a prior Dirichlet distribution with hyperparameter $\beta$. Then for each document $d$ the generative process assumes the following steps:

\begin{itemize}[labelindent=0.2em,labelsep=0.2em,leftmargin=*,itemsep=2pt,parsep=2pt]
\item From a collection wide Dirichlet distribution with hyperparameter $\alpha$, LDA draws a multinomial distribution $\theta_{d}$: $\theta_{d} \sim Dir.(\alpha)$.
\item For each word position $n=1,2,3,...,N_d$ in document $d$, LDA assigns a topic indicator $z_n$ by drawing topic from $\theta_d$: $z_n \sim Multi.(\theta_d)$. 
\item Using the assigned topic indicator ($z_n=k$) it then draws the actual word in position $n$ from the topic specific distribution over words: $w_n \sim Multi.(\varphi_{z_{n}})$.
\end{itemize}
In the LDA model the above process is repeated for each document in the collection. 

\begin{figure*}
  \centering
  \includegraphics[width=\textwidth]{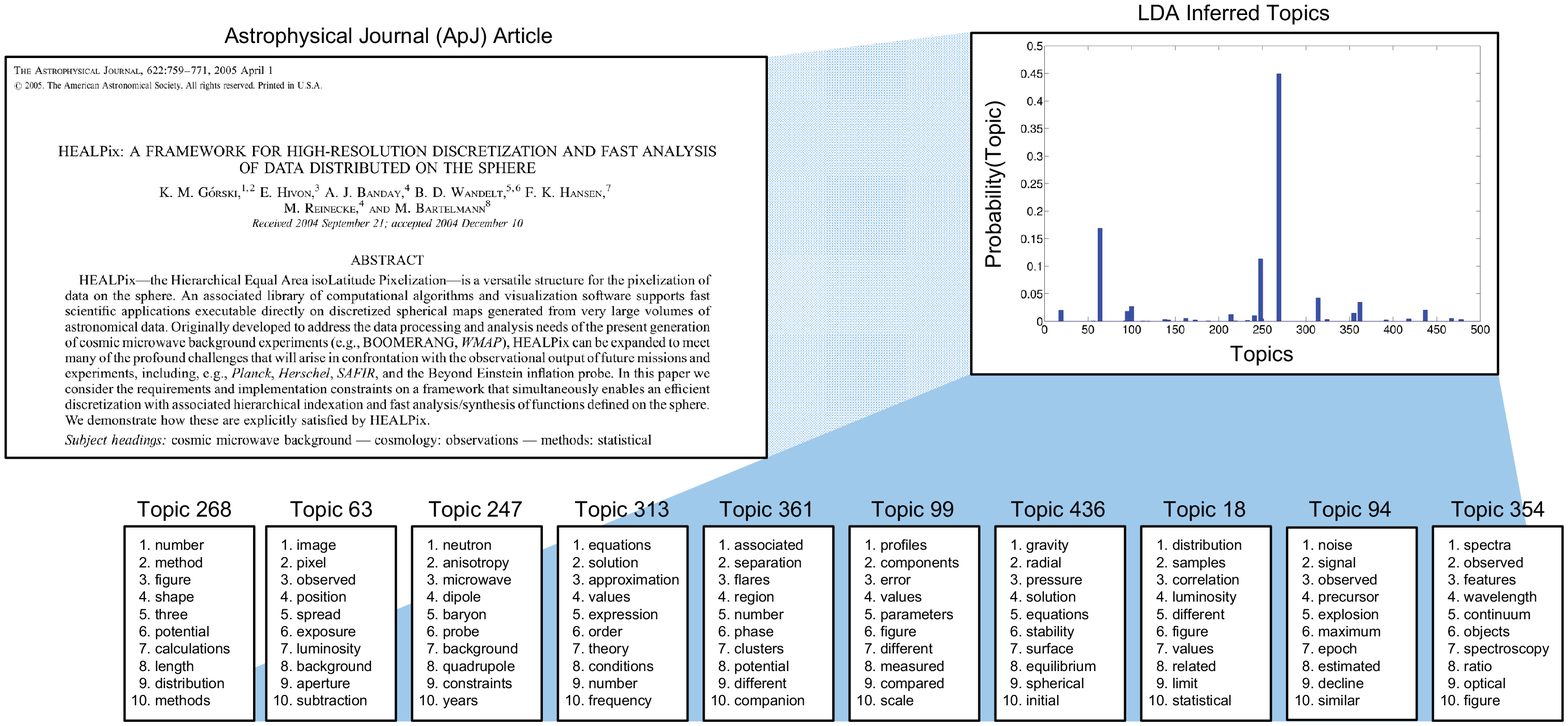}
  \caption{LDA representation of one of the most cited Astrophysical Journal (ApJ) article across all the articles published in the past 10 years. Shown in the upper right corner is the article's inferred document-topic distribution while on the bottom are the top ten most probable topics in the article. For each topic we show the top ten most probable words.}
  \label{fig:apj_lda_example}
\end{figure*}

Figure~\ref{fig:apj_lda_example} presents an example of representing an ApJ article using LDA with 500 topics. In the upper right corner we see the inferred document-topic multinomial distribution across the whole article. On the bottom are the inferred topic-word distributions for each of the ten most probable topics in the article. More specifically, for each topic-word distribution we show the top ten most probable words for that topic. Words are ranked based on their topic specific probability.

Once we represent all the ApJ articles in our collection as multinomial distribution over topics, the task of finding topically similar articles given an article that the user is currently viewing is formulated as finding similar probability distributions. In the probability simplex similarity between probability distributions is computed using information-theoretic measures such as: Kullback-Leibler (KL) or Jensen-Shannon (JS) divergence (JS is a symmetric form of KL) and Hellinger (He) distance. While computing similarity across all-pairs of articles is practically infeasible due to the time complexity of  $O(N^2)$, recent work \cite{Krstovski_ea_2013} has shown that tasking the problem as a nearest-neighbor (NN) search problem and by transforming divergences a substantially more efficient approach could be used for finding topically similar articles.

\subsection{Graphical Representation using Plate Notation}
\label{sec:2_2}
LDA is most conveniently represented using the probabilistic graphical model representation. This representation type depicts probability variables as nodes or vertices in the graph while the probabilistic relationship between the variables is represented using edges or arcs. It helps visualize the complex relationship across variables in latent variable models of text with a compact representation. 

\begin{figure*}
  \centering
  \includegraphics[width=0.4\textwidth]{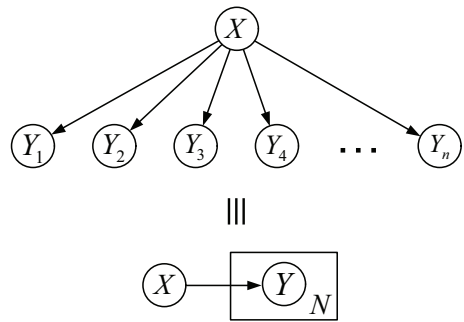}
  \caption{Example graphical model representation of a joint probability distribution $P(X,Y)$ and its plate notation equivalent.}
  \label{fig:plate_notation}
\end{figure*}

When models contain large set of variables its graphical representation could easily become cluttered and difficult to interpret. For example, variables are sometimes generated by independent draws from the same probability distribution. In such instances the graphical representation of the model would need to contain a node for each independent draw. The plate notation avoids the difficulty in interpreting such graphs by: (1) substituting the repeated nodes using a rectangle i.e. "plate"; (2) drawing a single node inside the plate and (3) annotating the plate with the number of repeated nodes found in the original graph. Figure~\ref{fig:plate_notation} shows an example graphical model representation of the joint probability $P(X,Y) = P(X)\prod_{i=1}^{N}P(Y_i|X)$ and its graphical representation equivalent using plate notation. A graphical representation of a model which is generated from its nested version and which contains all of its nodes is known as the "unrolled" version of the model. 

\begin{figure*}
  \centering
  \includegraphics[width=0.7\textwidth]{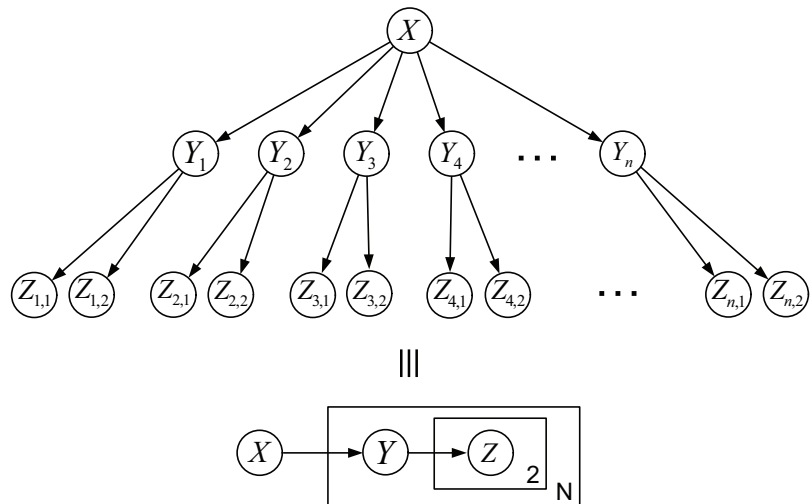}
  \caption{Nested plate notation representation of a joint distribution of three probability variables, $P(X,Y,Z) = P(X)\prod_{j=1}^{2}\sum_{i=1}^{N}P(Z_{i,j}|Y_i)P(Y_i|X)$. Shown on the bottom is its unrolled graphical model equivalent.}
  \label{fig:nested_plate_notation}
\end{figure*}

In many latent variable models of text, such as LDA, it is often the case to have a repeated set of nodes occurring on multiple levels. An example of such a graphical model is shown in Figure~\ref{fig:nested_plate_notation}. In this graph we have a joint distribution of three probability variables, $P(X,Y,Z) = P(X)\prod_{j=1}^{2}\sum_{i=1}^{N}P(Z_{i,j}|Y_i)P(Y_i|X)$. In instances like this, where there is a hierarchical relationship between the variables in the model, a more compact representation could be achieved by having nested plates. This notation is referred to as the "nested plate" and it is shown on the bottom of the figure.

\begin{figure*}
\centering
\includegraphics[width=0.40\columnwidth]{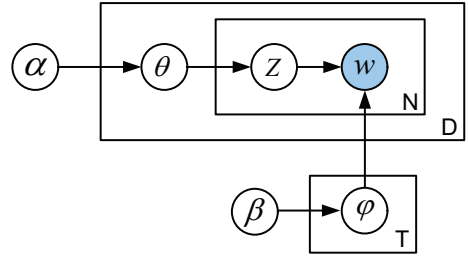}
\caption{Graphical representation of LDA using nested plate notation.}
\label{fig:lda}
\end{figure*}

Using the nested plate notation in figure~\ref{fig:lda} we show the graphical model of LDA.

\section{Multilingual Topic Models}
\label{sec:3}

While vocabulary mismatch occurs across scientific article from the same or different fields written in the same language, a more extreme version of mismatch occurs when articles are written in different languages. Documents that are translations of each other convey the same set of topics while their vocabularies are completely different. To model the generative process of documents that are translations of each other extensions of LDA have been proposed. These extensions jointly model the generative process of the translation documents using language specific vocabularies. For example the polylingual topic model (PLTM) \cite{Mimno_ea_2009}, which is the MLTM instance which we will use in this work, assigns a single document-topic distribution to a tuple of documents that are translations of each other and for each topic in the collection it assigns language specific topic-word distributions. Representing documents written in different language in a shared topic space facilitates the analysis of document relationships across languages. Such analysis allow, for example, to search documents that are translations of each other \cite{Krstovski_Smith_2013}. In other words, given a query document written in one language, with the MLTM based representation of a multilingual collection we would be able to find topically similar documents written in different languages. Aside from the benefits that it offers on the task of detecting document translation pairs, MLTM for example was shown to provide means for extracting parallel sentences from comparable corpora \cite{Krstovski_Smith_2016}.

\begin{figure}
  \centering
  \includegraphics[width=0.9\textwidth]{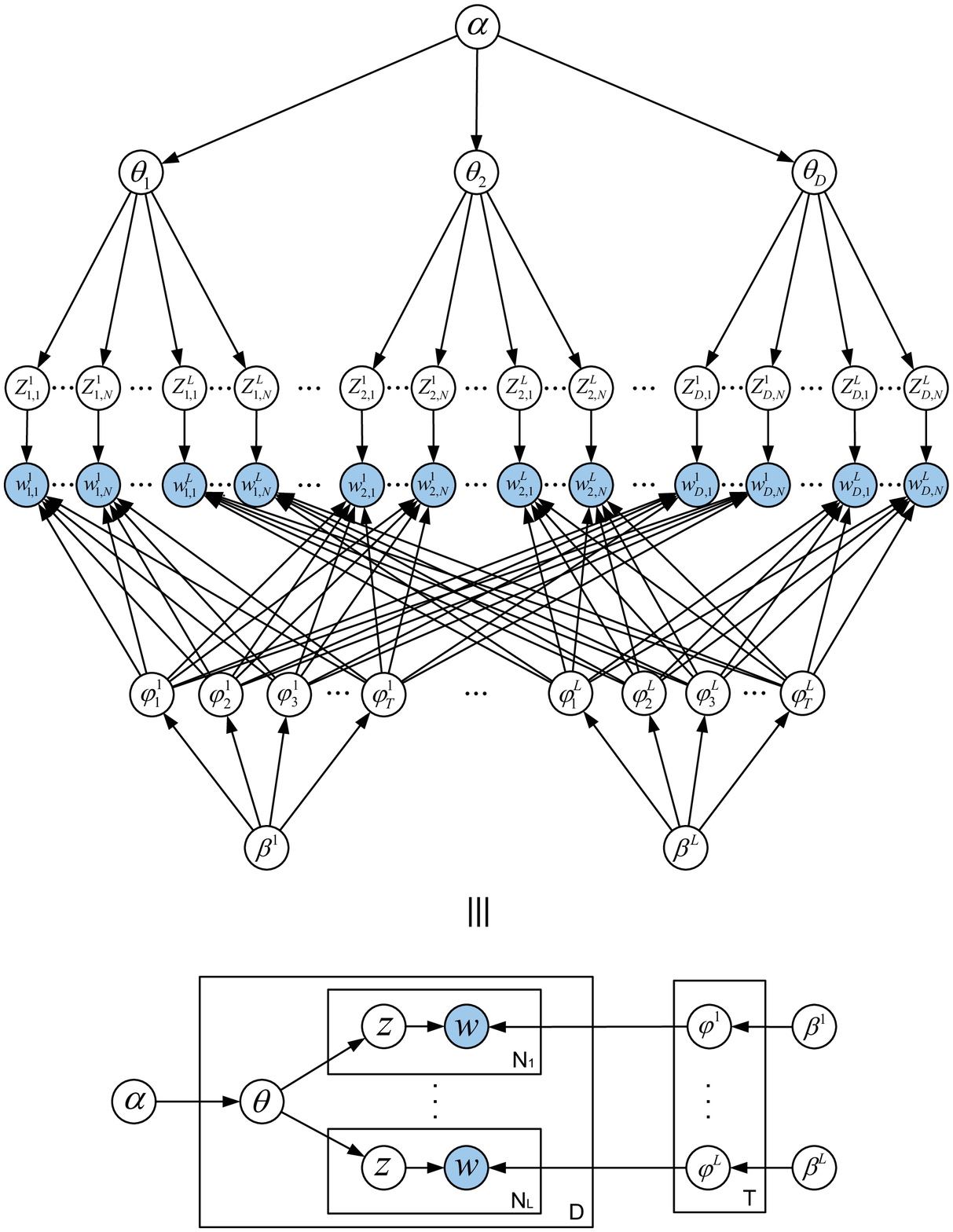}
  \caption{Unrolled graphical representation of the multilingual topic model (MLTM) with its nested plate notation equivalent displayed on the bottom.}
  \label{fig:pltm}
\end{figure}

For a collection of $d$ document tuples where each tuple consists of one or many documents in different languages $l=1,2,...,L$ that are topically similar $d$ = ($doc_1$, $doc_2$, $doc_3$, ..., $doc_L$),  MLTM assumes that documents in tuple $d$, that are written in different languages $L$, cover the same set of topics $\theta_{d}$. This is in contrast to LDA where we deal with a single document $d$ and a single document-topic distribution $\theta_{d}$. Figure~\ref{fig:pltm} shows the unrolled graphical model representation of MLTM. Shown on the bottom is the equivalent nested plate representation.

MLTM also assumes that each language has its own set of $K$ topic-word distributions $\varphi_{k}^{l}$ over the words in the language vocabulary $V_l$. In case of MLTM words in each language are drawn from language specific topic distributions $\varphi^l$: $w^l \sim p(w^l|z^l,\varphi^l)$. In addition, in MLTM topic assignments over words are drawn from tuple specific topic distributions $\theta_{d}$: $z^l \sim p(z^l|\theta_{d})$. 

What follows is a detailed description of the generative process that MLTM assumes and models. On a collection level, MLTM first generates a set of $k\in\{1,2,...,K\}$ topic-word distributions, $\varphi^{l}_{k}$ which are drawn from a Dirichlet prior with language specific hyperparameter $\beta^l$: $\varphi^l_k \sim Dir.(\beta^l)$. For each document $d^l$ in tuple $d$, MLTM then assumes the following generative process:

\begin{itemize}[labelindent=0.2em,labelsep=0.2em,leftmargin=*,itemsep=2pt,parsep=2pt]
\item Draw a tuple specific multinomial distribution $\theta_{d}$: $\theta_d \sim Dir.(\alpha_d)$.
\item For each language $l$ in document tuple $d$:
\begin{itemize}
\item For each word $w$ in document $d^l$:
\begin{itemize}
\item Choose a topic assignment $z_w \sim Multi.(\theta_{d})$
\item Choose a word $w \sim Multi.(\varphi_z^l)$
\end{itemize}
\end{itemize}
\end{itemize}

The model first draws a tuple specific distribution over topics $\theta_d$. As in the case with LDA, this distribution is drawn from a Dirichlet prior with hyperparameter $\alpha_d$: $\theta_d \sim Dir.(\alpha_d)$. 

Going over each language $l$ in the tuple, the model generates the $N^l$ document specific words by first drawing a topic assignment $z_w$ which is then used to select the language specific topic distribution over words $\varphi^l_z$. The actual word $w$ is drawn from the chosen topic-word distribution. 

\subsection{Inference in MLTMs}
\label{sec:3.1}
As in the case with LDA and other Bayesian models, computing the posterior distribution is intractable and therefore approximate approaches are used instead. When inferring posterior distributions with approximate methods researchers often utilize two approaches: Gibbs sampling and variational Bayes (VB). Most widely used of the two is the Gibbs sampling approach. The problem of approximating posteriors with Gibbs sampling is formulated as a sampling task while VB formulates the problem as an optimization task. What follows is a brief description of both approaches with emphasis on the important steps used in approximating the posterior per document-topic and per topic-word distributions.

\subsection{Gibbs Sampling}
\label{sec:3.2}
Gibbs sampling is a variant of the Markov chain Monte Carlo (MCMC) method  which constructs a Markov chain whose states are parameter settings and whose stationary distribution is the true posterior over those parameters. Rather than estimating the posterior document-topic $\theta_d$ and the topic-word $\varphi_t$ distributions, Gibbs sampling first estimates the posterior topic assignments $z$ which are then used to approximate $\theta_d$ and $\varphi_t$. Incorporating the multilingual concept in the Gibbs sampling inference approach doesn't imply increased parameter complexity and in fact is fairly straightforward. Approximating the tuple-topic distribution $\hat{\theta}_{d}$ is performed using the counts of the number of times topic $t$ was assigned in all documents of tuple $d$: $C^{DT}_{dt}$. Across the different languages in the collection we approximate the topic-word distributions using language specific matrices of counts $C^{W_{l}T}_{v_{l}t}$. The estimates of the two types of distributions are shown below:

\begin{eqnarray}
\label{eq:gibbs_theta_pltm}
  \hat{\theta}_{dt}=\frac{C^{DT}_{dt}+\alpha}{\sum^{T}_{j=1}C^{DT}_{dt_j}+T \alpha}
\end{eqnarray}

\begin{eqnarray}
\label{eq:gibbs_phi_pltm}
  \hat{\varphi}^{l}_{tw}=\frac{C^{W^{l}T}_{w^{l}t}+\beta^l}{\sum^{W_{l}}_{i=1}C^{W^{l}T}_{w^l_{i}t}+W^{l} \beta^{l}}
\end{eqnarray}

\subsection{Variational Bayes}
\label{sec:3.3}
\begin{figure*}
\centering
\includegraphics[width=0.55\textwidth]{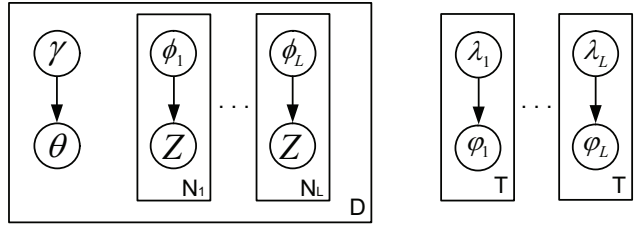}
\caption{Graphical model representation of the free variational parameters for the online variational Bayes approximation of the MLTM posterior.}
\label{fig:pltm_onlinevb}
\end{figure*}

The VB approach \cite{Jordan_ea_1999} defines the problem of approximating the posterior distributions as an optimization task. VB uses a family of probability distributions with variational parameters that simplifies the complex dependence of the latent variables $\theta$, $z$ and $\varphi^{l}$ in the MLTM. Their original dependence is broken down into dependencies of the individual model variables over the variational distributions $\gamma$, $\phi$ and $\lambda^l$ respectively. Figure~\ref{fig:pltm_onlinevb} shows the variational version of the MLTM and the free parameters used in our approach. In VB inference update steps for the variational parameters take into account the language $l$ count statistics as well as the topic-word distributions:

\begin{eqnarray}
\gamma_{dt}=\alpha+\sum_{l=1}^{L}\sum_{w=1}^{W^l}\phi^{dl}_{wt}\:n^{dl}_{w}\\
\phi^{dl}_{wt} \propto \exp\left\{E_{q}\left[\log\theta_{dt}\right]+E_{q}\left[\log\varphi^{dl}_{tw}\right]\right\}\\
\lambda^{l}_{tw} = \beta^l+\sum_{d=1}^{D}n^{dl}_{w}\phi^{dl}_{wt} 
\label{eq:pltm_gamma_phi}
\end{eqnarray}	

\subsection{Inferring Topics in Document Collections using LDA and MLTM}
\label{sec:3.4}
LDA and many of its variants, including MLTM, by another taxonomy are know as unsupervised models of text which means that the model doesn't rely on labeled data. This is in contrast to supervised models which are learned using a set of pairs of input data and labeled output values. When dealing with unsupervised models it is often the case that certain model parameters are inferred on one collection which is usually considered as the training step. Inferred values are then held fixed and are used to infer other set of model parameters onto a new or unseen collection. This step is usually referred as the test step. In case of LDA and MLTM, in the training step, the per topic-word distributions and the Dirichlet hyperparameters are inferred. Holding these values fixed, in the test step we infer per document-topic distributions on a set of unseen documents.

When using LDA and MLTM to represent document collections in the topic space one must first decide on the vocabulary that will be used to represent the documents. This vocabulary is often referred to as the effective vocabulary. Effective vocabulary is usually created by first running tf--idf statistics over all the tokens in the collection. In addition to removing the language specific set of stop words, the top most frequent words are treated as stop words and also removed. Words whose frequency of occurrence across the whole collection is low are also removed. It is typically the case that tokens whose frequency is less than 50, 25 or 10 are removed. Numeric tokens and tokens whose character length is less than four are removed as well. This filtering process creates a collection specific effective vocabulary. 

Since LDA and MLTM are bag of words models the ordering of words in the documents is not important while the document specific word frequency is. Therefore for efficiency, documents are represented as set of tuples where each tuple contains the integer representation of the word (by assigning each word in the effective vocabulary with an integer) along with the document frequency. 

\section{Using Multilingual Topic Models to Evaluate Representational Systems}
\label{sec:4}

\begin{figure*}
\centering
\includegraphics[width=1\textwidth]{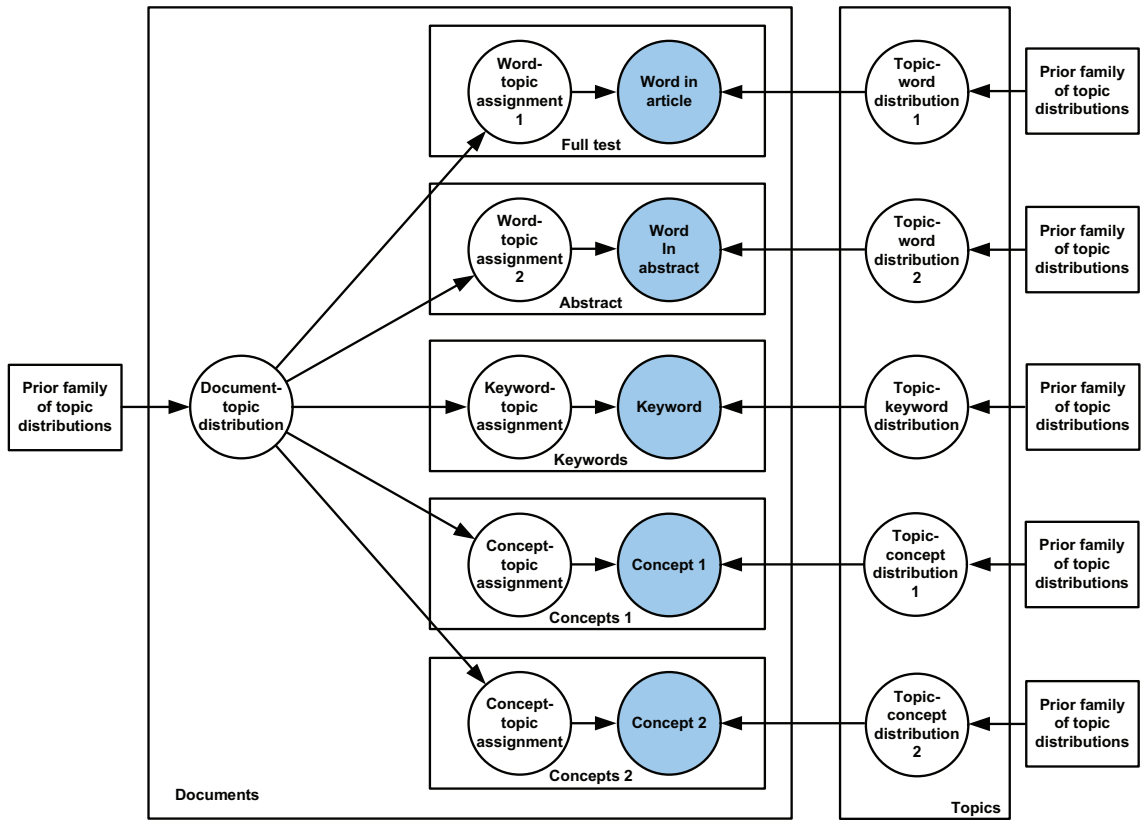}
\caption{Graphical model representation of the MLTM used to evaluation four different document representations. Each document representation in the MLTM model is treated as a translation of the original document in a different language.}
\label{fig:pltm_concept}
\end{figure*}

In this section we present our approach of using MLTMs to evaluate different representational systems. More specifically, we use MLTM to infer topics over articles represented with different approaches which we treat as translations of the original article in different languages. Figure~\ref{fig:pltm_concept} presents the graphical plate notation of the MLTM model that we use in our evaluation approach. With MLTM various article representations could be treated as a translation of the original article in a different language. In this particular MLTM we use and experiment with five representations: (1) original article, (2) abstract only, (3) keywords generated by the article author, (4) concept system 1 and (5) concept system 2. 

Once all the representations (including the original article representation) are mapped into the shared topic space we than setup an IR task. For a set of $D$ test tuples we have each tuple containing $l$ different representations of the article, $l=1,2,...,L$, where $l=1$ is the original article representation. In our case the original articles representation consists of the title, abstract, author affiliations and the remaining article sections. In our IR evaluation setup we treat the inferred per document-topic distributions $\theta^{l=1}$ over the original articles as queries $q = \theta^{l=1}_1, \theta^{l=1}_2, \theta^{l=1}_3,...,\theta^{l=1}_d$. For each of the $D$ query distributions we ran $L-1$ experiments where in each experiment the goal is to retrieve the most similar topic distribution across the set of $D$ distributions inferred over the $l$-th article representation where $l\neq1$. In the probability simplex similarity between topic distributions is performed using information-theoretic measures such as Kullback-Leibler (KL) and Jensen-Shannon (JS) divergence and Hellinger distance. In our experiments we use JS divergence which is the symmetric form of KL:
\begin{eqnarray}
KL = \sum_{i=1}^n p(x_i)\log\frac{p(x_i)}{q(x_i)} \\
JS = \frac{1}{2} \textrm{KL}\left(p, \frac{p+q}{2}\right) + \frac{1}{2} \textrm{KL}\left(q, \frac{p+ q}{2} \right)
\end{eqnarray}

Once similarity is computed between the query topic distribution and the $D$ topic distributions of the $l$ representation, documents are sorted. Our evaluation approach is based on the assumption that regardless of the approach used, document representation systems should be able to convey to the reader the same or very similar set of topics. Reflecting on the topical similarity between the original article and one of its representations, good representations should be topically most similar compared to others which means that they should always be at the top rank. We use precision of the top rank (P@1) to evaluate the retrieval performance of the document representation. 

\subsection{Experimental Setup}
\label{sec:4_1}
We showcase the ability to evaluate document representations with MLTMs using a collection of ApJ articles that were published in the time period between 2003 and 2013. These articles were obtained from the SAO/NASA Astrophysics Data System (ADS) \cite{Kurtz_ea_2005}. The collection consists of 32,393 such articles. For each article we used 5 different representations where the default representation consisted of all article sections, including title, author affiliation and abstract. In addition we use the abstract only, keyword only and two concept based representations which we detail in the next section. The ApJ collection was randomly split in a training and test sets where the training set contains 29,154 (90\%) and the test set 3,239 (10\%) of the articles. Table~\ref{tab:vocab_size} shows the size of the effective vocabulary used across the different document representations. We use the Mallet \cite{Mccallum_2002} implementation of MLTM.

\begin{table}
\centering
\begin{tabular}{|l| r|}
\hline
\hline
Representation Type & Vocabulary Size \\ [0.5ex]
\hline
\hline
Original article    & 171,419 \\
\hline
Abstract            & 18,159 \\
\hline
Keywords            & 8,385  \\
\hline
Concepts 1          & 8,619  \\
\hline
Concepts 2          & 1,579  \\
\hline
\end{tabular}
\caption{Size of the effective vocabulary across different article representations.}
\label{tab:vocab_size}
\end{table}

\subsection{Representing Documents Using Controlled Vocabulary}
\label{sec:4_2}
When representing documents using controlled vocabulary one may choose between keyword and concept based representation. While both approaches utilize a predetermined list of words to represent the document, their fundamental difference is in the underlying process that is used to generate them. Keywords are usually generated by humans and in most cases the author of the article is in charge of assigning them. Concepts on the other hand are generated through an automatic process that often involves analyzing documents and mining their content. Using a certain algorithmic approach or heuristics words and phrases are extracted from the article and are then used to represent it. In our experiments we used the author assigned keywords which for the Astrophysical journal are selected from a predetermined list of words. For the concept based representation of articles we used the ScienceWise \cite{Astafiev_ea_2012} and the most recently introduced Unified Astronomy Thesaurus (UAT) \cite{Accomazzi_ea_2014} system.

\subsection{Results}
\label{sec:4_3}
In Table~\ref{tab:p1} we show P@1 numbers across the four different article representations in our experimental setup across three different PLMT topic configurations T=50, 100 and 500. For the obtained results we observe that the ScienceWise system (Concepts 1) provides the best overall retrieval performance across the different topic configurations. This concept based representation provides better retrieval ability than the article's own abstract based representation. Across the four different representations, the article representation using keywords gives the overall worst performance. Observing the vocabulary size in Table~\ref{tab:vocab_size} we see that there is no correlation between the performance of the representation system and the size of the vocabulary used. 

\begin{table*}
\centering
\begin{tabular}{|l| r| r| r|}
\hline
\hline
Representation Type & T=50  & T=100 & T=500\\ [0.5ex]
\hline
\hline
Abstract            & 0.722  & 0.829  & 0.928 \\
\hline
Keywords            & 0.081  & 0.117  & 0.152 \\
\hline
Concept 1          & 0.766  & 0.893  & 0.970  \\
\hline
Concept 2          & 0.295  & 0.398  & 0.579  \\
\hline
\end{tabular}
\caption{Evaluating document representations using MLTM. P@1 computed across MLTM topic configurations with T=50, 100, 500 and 1000 topics.}
\label{tab:p1}
\end{table*}

\subsection{Determining the Optimal Training Set Size}
\label{sec:4_4}
In many instances, especially when dealing with new document collections, the number of articles processing by a concept based representational systems is very small. This is also the case with human annotated articles. While in the previous experimental setup we used 90\% of our collection as a training data in this experimental setup we wanted to observe the impact of the size of the training set on the variation in the performance of the different representational systems as measured by our approach. We divided the training set into 8 different subsets by removing additional 10\% of the original training set going down to using only 20\% of the overall collection to train our MLTM. Shown in Tables~\ref{tab:p50},~\ref{tab:p100} and~\ref{tab:p500} are the P@1 values across the different representational systems that we obtained in this process. Analysis shown in each table correspond to a different MLTM configuration with number of topics set to T=50, 100 and 500. Across the 3 tables we can make two observations. As we increase the number of topics we obtain better accuracy across all representation systems. This is the same observation that we made in Table~\ref{tab:p1} as well. More importantly we observe that across the different training set sizes the variation in performance (across all representation types) is not as significant as the variation across different topic configurations. This makes the MLTM representation type more practical for use  since models trained with smaller training sets give similar performances to models trained on large sets.

\begin{table*}
\centering
\begin{tabular}{|l| r| r| r| r| r| r| r| r|}
\hline
\hline
\multirow{2}{*}{Representation Type} & \multicolumn{8}{c|}{Training Size (\% of original collection)}\\[0.5ex]\cline{2-9} 
& 20 & 30 & 40 & 50 & 60 & 70 & 80 & 90 \\[0.5ex]
\hline
\hline
Abstract & 0.702 & 0.709 & 0.701 & 0.714 & 0.718 & 0.707 & 0.722 & 0.722 \\
\hline
Keywords & 0.077 & 0.094 & 0.086 & 0.088 & 0.085 & 0.092 & 0.089 & 0.081 \\
\hline
Concept 1 & 0.774 & 0.761 & 0.767 & 0.780 & 0.761 & 0.760 & 0.774 & 0.766 \\
\hline
Concept 2 & 0.285 & 0.300 & 0.308 & 0.278 & 0.320 & 0.299 & 0.308 & 0.295 \\
\hline
\end{tabular}
\caption{Impact of the training set size on the performance of the different representational systems measured by P@1. MLTM was configured with 50 topics.}
\label{tab:p50}
\end{table*}

\begin{table*}
\centering
\begin{tabular}{|l| r| r| r| r| r| r| r| r|}
\hline
\hline
\multirow{2}{*}{Representation Type} & \multicolumn{8}{c|}{Training Size (\% of original collection)}\\[0.5ex]\cline{2-9} 
& 20 & 30 & 40 & 50 & 60 & 70 & 80 & 90 \\[0.5ex]
\hline
\hline
Abstract & 0.808 & 0.812 & 0.815 & 0.823 & 0.826 & 0.820 & 0.829 & 0.829 \\
\hline
Keywords & 0.115 & 0.113 & 0.123 & 0.126 & 0.120 & 0.118 & 0.117 & 0.117 \\
\hline
Concept 1 & 0.888 & 0.887 & 0.886 & 0.880 & 0.875 & 0.883 & 0.868 & 0.893 \\
\hline
Concept 2 & 0.384 & 0.398 & 0.386 & 0.404 & 0.382 & 0.383 & 0.391 & 0.398 \\
\hline
\end{tabular}
\caption{Impact of the training set size on the performance of the different representational systems measured by P@1. MLTM was configured with 100 topics.}
\label{tab:p100}
\end{table*}

\begin{table*}
\centering
\begin{tabular}{|l| r| r| r| r| r| r| r| r|}
\hline
\hline
\multirow{2}{*}{Representation Type} & \multicolumn{8}{c|}{Training Size (\% of original collection)}\\[0.5ex]\cline{2-9} 
& 20 & 30 & 40 & 50 & 60 & 70 & 80 & 90 \\[0.5ex]
\hline
\hline
Abstract & 0.926 & 0.926 & 0.930 & 0.928 & 0.928 & 0.934 & 0.936 & 0.928 \\
\hline
Keywords & 0.143 & 0.153 & 0.156 & 0.147 & 0.143 & 0.147 & 0.151 & 0.152 \\
\hline
Concept 1 & 0.970 & 0.973 & 0.976 & 0.977 & 0.973 & 0.972 & 0.973 & 0.970 \\
\hline
Concept 2 & 0.561 & 0.587 & 0.586 & 0.584 & 0.589 & 0.577 & 0.582 & 0.579 \\
\hline
\end{tabular}
\caption{Impact of the training set size on the performance of the different representational systems measured by P@1. MLTM was configured with 500 topics.}
\label{tab:p500}
\end{table*}

\section{Using MLTM to Improve Representational Systems}
\label{sec:5}
\begin{figure}
\centering
\includegraphics[width=0.55\textwidth]{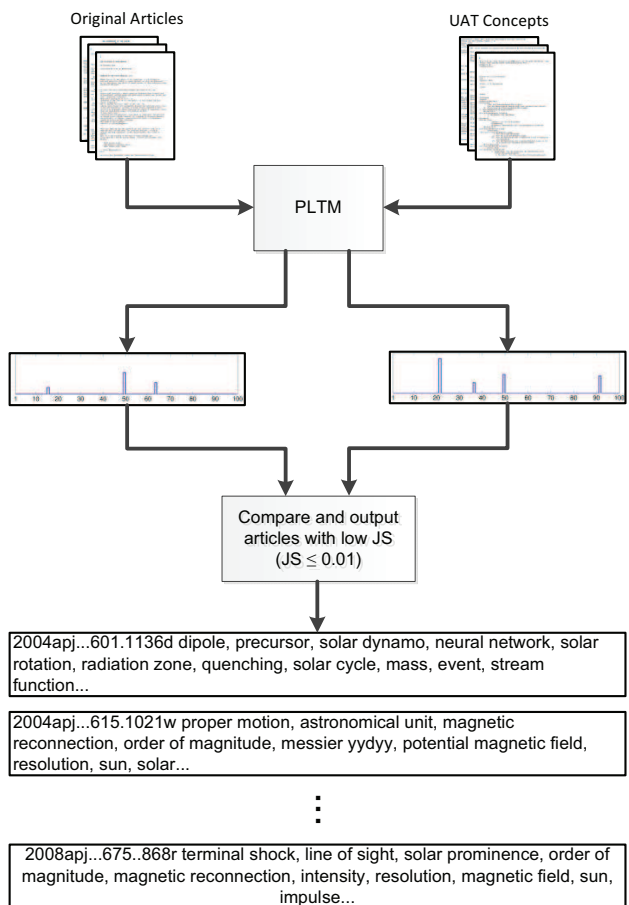}
\caption{Using MLTM to determine articles where concept based systems could provide better representation.}
\label{fig:pltm_evaluation}
\end{figure}

From the results presented in the previous section we observe that the UAT concept system offers far lower retrieval performance compared to ScienceWise. In this section we explain how the MLTM based approach of evaluating article representations could also help point out certain keywords and domains where the representation could be further improved. We use the UAT concepts as an example representation system. 
In the process of pointing out areas where representation could be further improved we start by looking into articles where the topic distributions across the two representations are not most topically similar. For this purpose we setup a JS divergence threshold (e.g. JS$\leq$0.01). The approach allows us to point out articles and determine a particular area or a sub-field where annotation could be improved by observing the author assigned keywords of the topically dissimilar representations. This in turn streamlines and navigates annotators through the process of assigning concepts/keywords. Figure~\ref{fig:pltm_evaluation} gives an outline of our approach. At the bottom we show example articles along with its keywords that did not pass the JS divergence threshold. Analyzing the author assigned keywords of the topically dissimilar article representations we discovered that articles published in the field of "solar physics" are not well annotated. To confirm our finding these articles were given to a team of librarians who manually verified that this is indeed the case and that the concept based representations of these articles could further be improved. 

\section{Conclusion}
\label{sec:6}
Representing articles using keyword and concept based representational systems offers a more functional and concise approach for determining articles that are similar to each other but are written using different vocabularies. For many collections and article types it is often the case that one needs to decide on the most suitable representation. To that end, in this paper we proposed a new approach for evaluating representational systems using MLTMs. Through a set of experiments on a collection of journal articles in the domain of Astrophysics we showed that our approach is able to clearly distinguish and rank different types of representational systems. Furthermore we showed that these analysis are invariant and could be performed using training sets of different sizes. The latter makes our approach very practical especially when working with small article collections. We also showed that our approach of treating different types of article representations as they were translations of each is capable of determining scientific sub-fields where a particular representational system could be improved.

\bibliography{mtm_kk}
\end{document}